# Data-Driven Bayesian Network Models of Hurricane Evacuation Decision Making

**Hui Sophie Wang[a*], Nutchanon Yongsatianchot[b], Stacy Marsella[c]**
[a]Institute for Experiential AI, Northeastern University, Boston, USA
[b]Faculty of Engineering, Thammasat School of Engineering, Thammasat University, Pathum thani, Thailand
[c]Khoury College of Computer Science & Department of Psychology, Northeastern University, Boston, USA

**Abstract:** Hurricanes cause significant economic and human costs, requiring individuals to make critical evacuation decisions under uncertainty and stress. To enhance the understanding of this decision-making process, we propose using Bayesian Networks (BNs) to model evacuation decisions during hurricanes. We collected questionnaire data from two significant hurricane events: Hurricane Harvey and Hurricane Irma. We employed a data-driven approach by first conducting variable selection using mutual information, followed by BN structure learning with two constraint-based algorithms. The robustness of the learned structures was enhanced by model averaging based on bootstrap resampling. We examined and compared the learned structures of both hurricanes, revealing potential causal relationships among key predictors of evacuation, including risk perception, information received from media, suggestions from family and friends, and neighbors evacuating. Our findings highlight the significant role of social influence, providing valuable insights into the process of evacuation decision-making. Our results demonstrate the applicability and effectiveness of data-driven BN modeling.

## 1. INTRODUCTION

Hurricanes lead to substantial economic and human costs [15]. During these events, individuals face difficult evacuation decisions amidst uncertainty and stress. Enhancing our understanding of this decision-making process is crucial for government policy making, where Artificial Intelligence (AI) and Machine Learning (ML) techniques can serve as valuable tools.

Empirical studies have identified various predictors of evacuation decision. The results of these studies are summarized in meta-analyses [1], [8]. Most empirical studies analyze these predictors independently in relation to evacuation decision without examining the relationships among the predictors. Therefore, despite the significant findings, these studies offer limited insight into the underlying mechanisms of the evacuation decision-making processes.

This leads to a consideration of two influential theoretical frameworks: Protection Motivation Theory (PMT) [17] and Protective Action Decision Model (PADM) [13], which provide deep insights into the mechanism of the decision-making processes in the context of natural hazards. While theoretical frameworks provide valuable insights, they operate at a level of abstraction that does not directly engage with empirical data. Computational models can serve as bridge between theoretical models and empirical data. They excel at capturing intricate and dynamic interconnections between variables. Through simulation and prediction, they provide a robust platform for testing/refining theories and hypotheses.

In this work, we propose Bayesian Network (BN) for modeling evacuation decision-making. A BN is a probabilistic graphical model that represents a set of variables and their dependencies via a directed acyclic graph [11]. It has been utilized in a wide range of applications, including risk assessment [6]. There are several advantages of using BN. Evacuation decision-making involves significant uncertainty, and BN's probabilistic approach is especially well-suited for reasoning under uncertainty. In addition, BN's graphical structure provides an intuitive means of representing complex causal influences. Its structure can be elicited from experts, learned from data, or a combination of both, giving BN the ease and flexibility to incorporate theories and domain knowledge. In sum, BN is a powerful statistical and machine learning model with high interpretability, making it well-suited for inference and learning tasks in high-stake decision making scenarios.

We conducted questionnaire studies of two significant hurricanes from the 2017 Atlantic hurricane season: Hurricane Harvey and Hurricane Irma. Given the differences between these hurricanes, we constructed a separate BN for each. We selected relevant variables for the BNs using mutual information before applying two constraint-based structure learning algorithms. Analyzing the results of the learned structures revealed potential causal relationships among key predictors of evacuation. Our findings demonstrate the applicability and effectiveness of data-driven BN modeling in uncovering complex interrelated factors influencing evacuation decision making.

## 2. RELATED WORK

With respect to empirical findings on the predictors of hurricane evacuation decision, Baker [1] identified several strong predictors of evacuation decisions, including location risk level, official notices, housing, risk perception, and storm features. More recently, Huang et al. [8] investigated the effect sizes of various predictors through statistical meta analysis, and found consistent significant predictors including official notices, mobile home, location risk level, observations of social and environmental cues, and expectations of severe personal impacts. They also found that reliance on information from peers, home ownership, expected hurricane intensity, and expected nearby landfall are weaker predictors, while expected wind damage, flood damage, and personal casualties have a moderately consistent and small significant correlation with evacuation decisions.

On the side of theoretical models, Protection Motivation Theory (PMT) [17] and Protective Action Decision Model (PADM) [13] are two major theoretical frameworks for understanding how people respond to hurricanes and other environmental hazards. PMT proposes that protection motivation, which in this context refers to whether to evacuate or not, is governed by two cognitive processes: threat appraisal and coping appraisal. PADM is based on research findings on people's responses to environmental hazards. It provides a detailed account of the decision-making process, encompassing three main stages: information processing, psychological decision processes, and behavioral response. The first stage includes the exposure to and processing of environmental cues, social cues, and warning messages. The second stage involves pre-decisional processes, core perceptions of environmental threat, protective actions and social stakeholders, as well as protective action decision making.

A number of statistical and computational models have been proposed for modeling evacuation decision during natural disasters [7], [9], [14], [19], [23], [26]. The most frequently used computational models in evacuation decision modeling are various forms of logistic regression [7], [14], [23]. Logistic regression models estimate the probability of evacuation using all available predictors, but they typically do not consider potential relationships among these predictors. Huang et al. [9] used Structural Equation Modeling (SEM) to investigate the relationships among predictors of evacuation decision, focusing on the mediation effects of key variables such as perceived storm characteristics, expected personal impacts, and perceived evacuation impediments. The structure of the model and the hypotheses are based on PADM. SEM integrates factor analysis and multiple regression to model complex causal relationships. However, SEM assumes the underlying relationships are linear. On the other hand, BN does not make assumptions about the type of the relationships or the distributions, and its probabilistic representation is particularly suitable for reasoning under uncertainty. Additionally, while BN can integrate theoretical and domain-specific knowledge, it can also be data-driven. This capability allows BN to build models based on data without strong initial hypotheses about the relationships between variables, thereby enabling knowledge discovery from data.

Recently, Partially Observable Markov Decision Processes (POMDPs) have also been utilized to model evacuation decision making [19], [26]. A POMDP is a generalization of the Markov Decision Process (MDP), which is a framework designed for sequential decision-making under uncertainty. While both MDPs and Bayesian Networks (BNs) are probabilistic graphical models, they serve different purposes. MDPs focus on determining an optimal sequence of actions to maximize the expected cumulative rewards over time. In contrast, BNs aim to model the relationships among variables. Despite these differences, it is possible to represent the state of an MDP using a BN, leading to the creation of a Dynamic Bayesian Network (DBN) or a Dynamic Influence Diagram (DID) that integrates with the MDP framework, as demonstrated by Sankar et al.[19]. However, there are several limitations to their work. First, the structure of the BN was manually defined without taking into account existing theoretical frameworks and empirical studies, and it was not subsequently evaluated against the collected data. A large portion of the parameters for the BN were also defined manually. Additionally, they conducted a survey on multiple hurricanes, pooling data from two very different hurricanes without providing any justification.

## 3. DATA

We conducted separate questionnaire studies on Amazon Mechanical Turk (MTurk) for two major hurricanes from the 2017 Atlantic hurricane season: Harvey and Irma. Hurricane Harvey was a devastating Category 4 hurricane that made landfall on Texas and Louisiana in the U.S. in August 2017. The questionnaire on Harvey was active on MTurk from October 30th to November 27th, 2017, targeting residents of Texas and Louisiana. We obtained 508 responses after removing duplicated MTurk worker IDs. Hurricane Irma was a catastrophic Category 5 hurricane that made a total of seven landfalls in September 2017, causing significant damage across the Caribbean and the southeastern U.S. The questionnaire on Irma was active from December 17th, 2017 to February 20th, 2018, targeting residents of Florida. 1,180 responses were obtained for Irma. It should be noted that both survey samples contain a higher proportion of females and younger individuals compared to the 2016 American Community Survey census data in the targeted regions [24].

The design of the questionnaire was informed by theoretical frameworks such as PMT and PADM, as well as previous empirical studies on hurricane [5], [8], [12]. The variables assessed by the questionnaire can be grouped into eight categories: demographic information, characteristics of the house, information received from media, social influences, evacuation notices from officials, evacuation difficulties, risk perception, and evacuation decision. Risk perception corresponds to threat appraisal in PMT and to the core perception of environmental threats in PADM. The questions included perceived wind risk to safety and property, perceived flood risk to safety and property, and risk of stay. Evacuation difficulties correspond to coping appraisal in PMT and to situational impediments in PADM. We considered the following difficulties: having children, having elders, having family members with special needs, having pets, evacuation expense, no transportation, no place to go and job obligations. Regarding the characteristics of the house, we asked about its type, material, ownership, insurance status, and distance to the coast. We asked participants about how much they saw the following information on TV/radio and social media, including images or videos showing damage caused by the storm, the strength of the storm, people making preparations, people staying, people leaving, and traffic conditions. Social influence questions included suggestions from family and friends, as well as the behaviors of neighbors. We also asked participants whether they had received any evacuation notices from officials.

## 4. APPROACH

Our goal is to develop and compare two Bayesian Network models using questionnaire data from Hurricanes Harvey and Irma. Although predicting evacuation is important, this work focuses on knowledge discovery, aimed at identifying the graphical structure that best represents the relationships among the variables influencing evacuation decision.

Learning BN structure from data is a very challenging problem. The number of possible structures for a BN grows super-exponentially with the number of variables, making structure learning an NP-hard problem [10]. In addition, the true structure may not be identifiable from data. There can be many equally good BN structures that cannot be distinguished based solely on data [11]. Moreover, the collected data can be noisy and biased, potentially leading to spurious edges. Despite these challenges, BN has been applied to many real-world problems and was shown to be effective in discovering both known and previously unknown relationships [4], [18], [22]. In this section, we describe our approach, which consists of feature selection and structure learning.

### 4.1. Variable Selection

Performing variable selection before structure learning offers several advantages. It significantly reduces the solution space, allows us to focus on the most relevant variables, and enables closer examination and better interpretation of the learned structure. Additionally, this approach helps reduce overfitting and can lead to a more robust structure with fewer spurious edges.

We use mutual information (MI) [2] to select variables that are most relevant to evacuation. MI quantifies how much the uncertainty of a variable is reduced by observing the value of another variable. It is commonly utilized as a filter method for feature selection in machine learning. Before variable selection, we conducted preprocessing, where variables with a high number of levels were grouped into categories. As a result, most variables were reduced to 2 or 3 levels. Because all the risk perception variables were highly correlated with each other, we retained only the risk of stay. We computed MI between feature variables and evacuation, selecting those variables whose MI exceeds 1% of the entropy of evacuation. This procedure results in 12 feature variables for the Harvey dataset and 10 for the Irma dataset. Figure 1 shows the MI of all the features ranked in descending order.

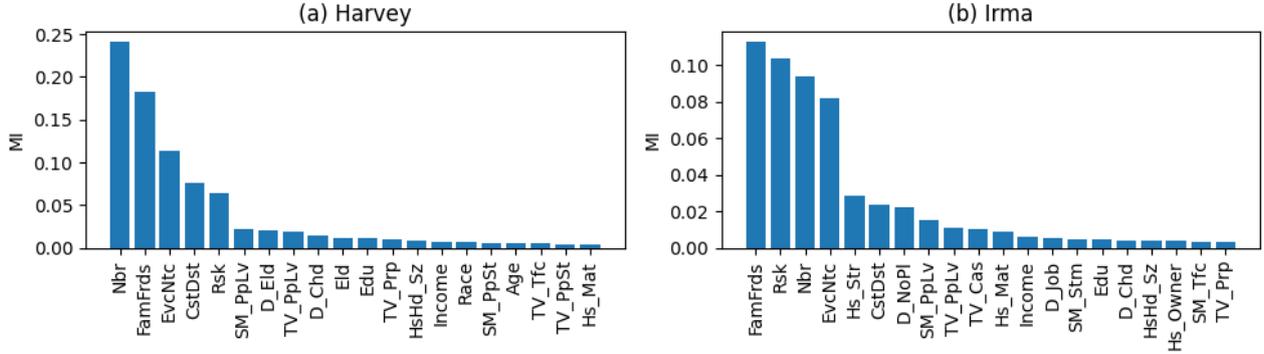

Figure 1. Mutual Information between feature variables and evacuation (top 20).
CstDst: distance to coast; D_Chd: difficulty involving children; D_Eld: difficulty involving elders; D_Job: difficulty job obligation; D_NoPl: difficulty no place to go; Edu: Education; Eld: has elders in household; EvcNtc: receiving evacuation notice; FamFrds: evacuation suggestions from family and friends; Hshd_Sz: household size; Hs_Mat: house material; Hs_Owner: house owner; Hs_Str: house structure; Income: household income; Nbr: neighbors evacuating; Rsk: perceived risk of staying; SM_PpLv: people leaving on social media; SM_Ppst: people staying on social media; SM_Stm: social media storm strength; SM_Tfc: traffic on social media; TV_Cas: casualty on TV; TV_PpLv: people leaving on TV; TV_Ppst: people staying on TV; TV_Prp: people preparing on TV; TV_Tfc: traffic on TV.

**4.2. Structure Learning**

A large number of BN structure learning algorithms have been proposed, and they generally fall into three categories: constraint-based, score-based, and hybrid. Constraint-based algorithms use conditional independence (CI) tests to find an equivalent class of networks that are consistent with data. Score-based algorithms use search and optimization to find the highest-scoring network structure based on a scoring criterion. Hybrid algorithms combine constraint-based and score-based approaches. For a comprehensive review of BN structure learning algorithms, refer to [10].

There is no consensus about the best structure learning algorithm. The choice of an appropriate algorithm is contingent upon the specific objectives and requirements of the problem at hand. Both constraint-based and score-based approaches have distinct strengths and weaknesses that make them suitable for different scenarios. Constraint-based algorithms explicitly test for conditional independence with the aim of discovering causal relationships, assuming causal faithfulness and causal sufficiency [10]. Meanwhile, they are sensitive to the accuracy of the conditional independence (CI) tests, which can be unreliable when the sample size is small. Score-based algorithms evaluate the entire network structure at once, which can lead to a globally optimal structure. However, the learned edges may not represent causal relationships, because they are typically chosen based on maximizing the likelihood of the data rather than explicit tests of causal influence.

Given that knowledge discovery is the main focus of this work, we primarily consider constraint-based algorithms. We utilize 'bnlearn' [20], a well-established R package that supports structure learning, parameter estimation, and inference for BNs. We employ two constraint-based algorithms: PC-stable [3] and Inter-IAMB [25]. PC-stable is based on PC algorithm, a global discovery algorithm that attempts to learn the entire graph structure by first constructing the skeleton and then determining the edge orientations. Inter-IAMB is a variant of the Incremental Association Markov Blanket (IAMB) algorithm, which focuses on learning the local structure of each variable by utilizing the concept of a Markov blanket. By employing these two distinct constraint-based algorithms, we aim to gain a deeper understanding of the reliability of the learned structures by comparing their results. The CI test used is mutual information with a significance level of 0.05.

To aid the structure learning algorithms in determining the directionality of edges, certain edges were excluded from the network structure by employing a blacklist, in which we specify unlikely or irrelevant causal influence directions. For instance, edges from evacuation to any of the feature variables are prohibited. Moreover, the feature variables are grouped into three tiers. The first tier comprises demographic variables and house characteristics. The second tier encompasses information received from media, social influences, and evacuation notices. The third tier includes risk perception and evacuation difficulties, which are key appraisal variables in PMT. Edges from higher tiers to lower tiers are prohibited, as they are considered unlikely or irrelevant, while edges from lower tiers to higher tiers and those within the same tier are permitted.

Furthermore, to enhance the robustness of the learned structure, model averaging is employed [21]. We perform bootstrap resampling of the dataset 1,000 times to determine the frequency with which each edge appears in the learned structure. This frequency is used as a measure of the confidence of each edge.

## 5. RESULTS

The BN structures learned for Harvey and Irma are shown in Figure 2 and Figure 3, respectively. The edges shown in the graph structures are the ones with confidence above 0.3, which means they appear in at least 30\% of the learned structures based on bootstrap resampling. The frequencies and directionality of these edges are shown in Table 1 and Table 2. These edges are further grouped into three levels based on their frequency: high confidence ($> 0.5$), medium confidence (0.4-0.5), and low confidence (0.3-0.4). We consider edge directions to be unreliable when the frequency is below 0.6, which means both directions occur with about the same frequency. Nevertheless, when the frequency is above 0.6, its direction is not necessarily reliable. Constraint-based algorithms cannot distinguish between graphs that belong to the same I-equivalence class, and the randomly selected edge directions may not result in an equal split due to the constraint of the v-structures [11]. In this section, we will examine the learned structures in detail and provide interpretations and comparisons.

### 5.1. BN Structure learned from Harvey Data

The results of Harvey are summarized in Figure 2 and Table 1. First, we examine variables that exert a direct influence on evacuation decision. Both algorithms identified with high confidence that coast distance, neighbors evacuating, and evacuation suggestions from family and friends directly influence evacuation decision. The influence of evacuation notice on evacuation decision is less certain compared to the aforementioned variables. Notably, there is an absence of a direct relationship between risk perception and evacuation decision, contrary to the postulations of PMT and PADM. This absence is likely attributable to the lower MI of risk perception compared to the other four variables, as shown in Figure 1 (a).

Next, we focus on risk perception. Both algorithms concur with high confidence that risk perception is directly influenced by suggestions from family and friends. The influence of media information on risk perception is less certain, and this effect is observed only in the structure learned by Inter-IAMB. Inter-IAMB also identified an undirected relationship with low confidence between risk perception and evacuation difficulty involving children.

Suggestions from family and friends occupy a central role in the learned structures. Not only does it influence risk perception, but it also appears to influence the reception of certain information on social media, regarding people leaving. This might be because the communications with family and friends occurred on social media. Note that this edge direction occurred approximately 70% of the time, implying that the opposite direction is also a considerable possibility.

Furthermore, an association between the suggestions from family and friends and the evacuation behaviors of neighbors is identified with high confidence by both algorithms. This relationship is challenging to explain, as typically, there is no direct communication between one's family and friends and their neighbors, unless such interactions occur on social media. Further investigation is necessary to better understand the flow of information and communication on social media, as well as the communication channels individuals use with their family, friends, and neighbors.

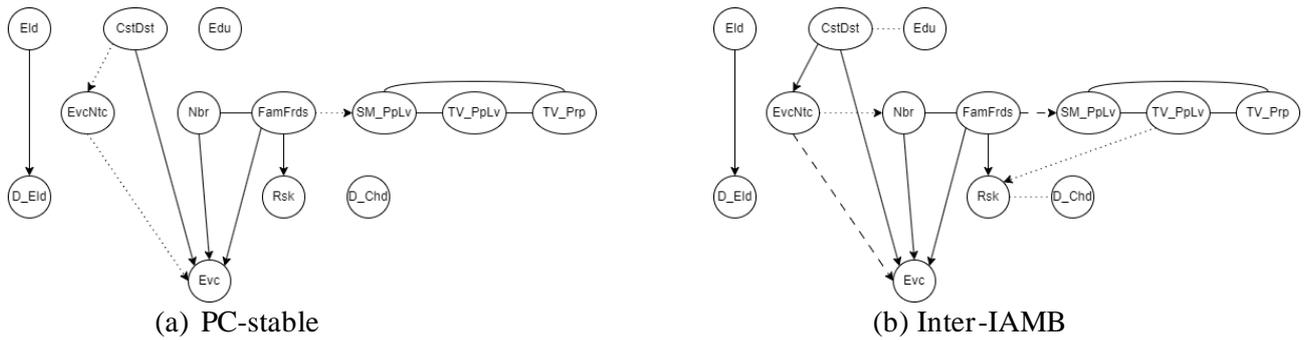

(a) PC-stable  (b) Inter-IAMB

Figure 2: BN Structure learned from Harvey Data. Solid arrow and line: high confidence; dashed arrow and line: medium confidence; dotted arrow and line: low confidence.

Table 1. Edge Frequency after Model Averaging (Harvey)
Left: PC-stable; Right: Inter-IAMB

| From | To | Frequency | Direction | From | To | Frequency | Direction |
|---|---|---|---|---|---|---|---|
| Eld | D_Eld | 1.000 | 1.000 | Eld | D_Eld | 1.000 | 1.000 |
| SM_PpLv | TV_PpLv | 1.000 | 0.560 | SM_PpLv | TV_PpLv | 1.000 | 0.504 |
| Nbr | Evc | 0.997 | 0.948 | Nbr | FamFrds | 0.998 | 0.529 |
| Nbr | FamFrds | 0.992 | 0.511 | Nbr | Evc | 0.997 | 0.970 |
| TV_Prp | TV_PpLv | 0.983 | 0.590 | TV_Prp | TV_PpLv | 0.991 | 0.563 |
| FamFrds | Rsk | 0.726 | 0.988 | FamFrds | Rsk | 0.805 | 0.985 |
| FamFrds | Evc | 0.691 | 0.927 | TV_Prp | SM_PpLv | 0.723 | 0.521 |
| CstDst | Evc | 0.649 | 1.000 | CstDst | Evc | 0.693 | 0.994 |
| TV_Prp | SM_PpLv | 0.526 | 0.589 | FamFrds | Evc | 0.683 | 0.964 |
| EvcNtc | Evc | 0.373 | 1.000 | CstDst | EvcNtc | 0.541 | 1.000 |
| CstDst | EvcNtc | 0.361 | 1.000 | FamFrds | SM_PpLv | 0.493 | 0.710 |
| FamFrds | SM_PpLv | 0.314 | 0.728 | EvcNtc | Evc | 0.418 | 0.993 |
| | | | | CstDst | Edu | 0.384 | 0.513 |
| | | | | TV_PpLv | Rsk | 0.318 | 0.987 |
| | | | | EvcNtc | Nbr | 0.313 | 0.685 |
| | | | | Rsk | D_Chd | 0.302 | 0.570 |

## 5.2. BN Structure learned from Irma Data

First, we examine the factors directly influencing the evacuation decision. Both algorithms converge on the following five direct influences: lack of a place to go, risk perception, suggestions from family and friends, receipt of an evacuation notice, and neighbors evacuating. The first three of these factors have very high confidence levels (above 0.8 in PC-stable and above 0.9 in Inter-IAMB). Note that risk perception and lack of a place to go correspond to key constructs in PMT and PADM. These findings thus provide support for PMT and PADM. The influence of official evacuation notice on evacuation decision is consistent with empirical findings [1], [8]. The influence of neighbors evacuating is also consistent with empirical findings [8]. Furthermore, it aligns with PADM, which views social cues as initiators of pre-decisional processes. However, the edge with relatively high confidence from it to the evacuation decision indicates a direct influence, rather than mediation through a series of other variables as postulated by PADM.

Next, we examine the factors influencing risk perception. Both algorithms identify two variables, suggestions from family and friends and information about people leaving on TV—as directly affecting risk perception, with higher confidence levels in Inter-IAMB compared to PC-stable. Additionally, Inter-IAMB identifies an influence of coastal distance on risk perception, which is reasonable, although it is inferred with relatively low confidence.

Furthermore, both algorithms concur with high confidence that coastal distance directly influences both the receipt of evacuation notice and neighbors' evacuation. This finding is reasonable and could explain the identified association between receiving evacuation notice and neighbors' evacuation. Additionally, associations between house structure and coastal distance, evacuation notices, and neighbors' evacuation were identified with medium or low confidence. This is likely due to the presence of mobile homes, although they represent only a small percentage (6.8%) of the total.

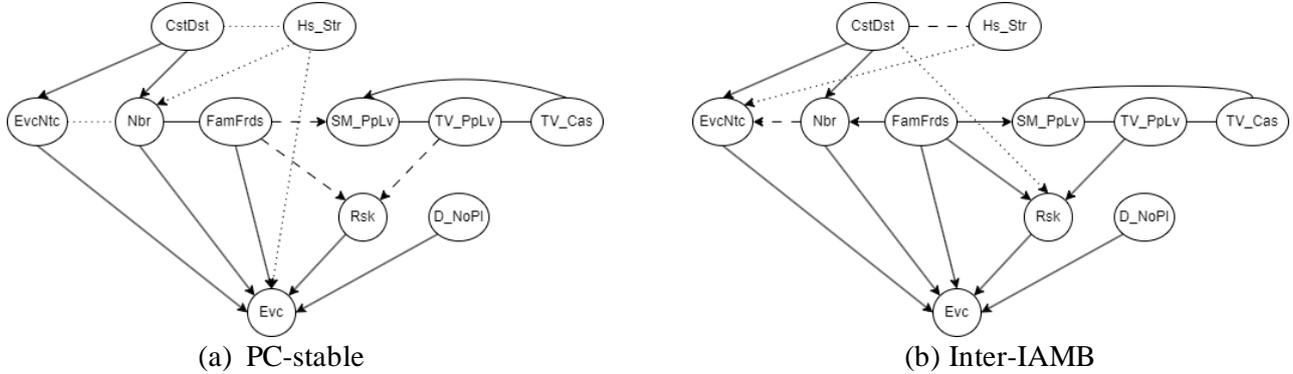

          (a) PC-stable                                       (b) Inter-IAMB
Figure 3: BN Structure learned from Irma Data. Solid arrow and line: high confidence; dashed arrow and line: medium confidence; dotted arrow and line: low confidence.

Table 2. Edge Frequency after Model Averaging (Irma)
Left: PC-stable; Right: Inter-IAMB

| From | To | Frequency | Direction | From | To | Frequency | Direction |
| --- | --- | --- | --- | --- | --- | --- | --- |
| D_NoPl | Evc | 0.978 | 1.000 | CstDst | EvcNtc | 1.000 | 0.954 |
| TV_PpLv | SM_PpLv | 0.947 | 0.542 | TV_PpLv | SM_PpLv | 1.000 | 0.543 |
| TV_Cas | TV_PpLv | 0.946 | 0.534 | D_NoPl | Evc | 0.999 | 1.000 |
| Nbr | FamFrds | 0.910 | 0.540 | Rsk | Evc | 0.998 | 1.000 |
| CstDst | EvcNtc | 0.876 | 0.969 | FamFrds | Nbr | 0.994 | 0.702 |
| FamFrds | Evc | 0.838 | 1.000 | TV_Cas | TV_PpLv | 0.986 | 0.532 |
| Rsk | Evc | 0.826 | 1.000 | FamFrds | Evc | 0.961 | 1.000 |
| EvcNtc | Evc | 0.665 | 1.000 | TV_PpLv | Rsk | 0.902 | 0.995 |
| TV_Cas | SM_PpLv | 0.614 | 0.643 | EvcNtc | Evc | 0.874 | 1.000 |
| CstDst | Nbr | 0.531 | 0.968 | FamFrds | Rsk | 0.814 | 0.998 |
| Nbr | Evc | 0.520 | 1.000 | FamFrds | SM_PpLv | 0.763 | 0.723 |
| FamFrds | Rsk | 0.489 | 0.994 | CstDst | Nbr | 0.753 | 0.994 |
| TV_PpLv | Rsk | 0.485 | 0.995 | Nbr | Evc | 0.601 | 1.000 |
| FamFrds | SM_PpLv | 0.463 | 0.645 | SM_PpLv | TV_Cas | 0.535 | 0.574 |
| EvcNtc | Nbr | 0.398 | 0.597 | Nbr | EvcNtc | 0.483 | 0.647 |
| CstDst | Hs_Str | 0.374 | 0.503 | CstDst | Hs_Str | 0.455 | 0.500 |
| Hs_Str | Evc | 0.351 | 1.000 | CstDst | Rsk | 0.320 | 0.997 |
| Hs_Str | Nbr | 0.303 | 0.941 | Hs_Str | EvcNtc | 0.311 | 0.854 |

## 5.3. Comparing the BN structures of Harvey and Irma

Several similarities can be observed in the learned structures of Harvey and Irma. Among the identified direct factors influencing evacuation decision, two of them are shared by both hurricanes: suggestions from family and friends and neighbors evacuating. This underscores the significant role of social influence in evacuation decisions. Moreover, these two factors are correlated, presenting an interesting relationship whose underlying mechanism requires further investigation.

As for factors that directly influence risk perception, suggestions from family and friends were identified with high confidence in both hurricanes. This highlights the importance of social influence on risk perception. On the other hand, the influence of media information on risk perception was identified with varying degrees of confidence. Specifically, information about people leaving on TV appears to be the one that influences risk perception, although all the media information variables are highly correlated with each other in both hurricanes. The social nature of this information further underscores the importance of social influence on risk perception.

It is worth noting that the reception of information about people leaving on social media appears to be influenced by suggestions from family and friends. This relationship was identified in both hurricanes with varying degrees of confidence. Although the direction of this influence is consistent in both hurricanes, the confidence in this direction is not very high, suggesting it is likely a correlation rather than causation. A potential common cause linking these two variables is the use of social media.

As for the differences between the learned structures of the two hurricanes, one major difference is that risk perception and evacuation difficulty were found to directly influence evacuation only in Irma, but not in Harvey. It is likely that the relationship exists in Harvey, but it was not strong enough compared to the influence of other variables. Additionally, the sample size for Harvey is much smaller than for Irma, which may make the structure learned from Harvey less reliable.

## 6. CONCLUSION

This work makes several important contributions to the understanding of evacuation decision-making during hurricanes. First, it demonstrates the applicability and effectiveness of BNs in modeling complex decision-making processes in the context of evacuations during natural disasters. By comparing BN structures learned from questionnaire data of two significant hurricanes, Harvey and Irma, this research identifies key predictors and their interrelationships, offering deeper insights beyond traditional regression models. Another key contribution of this work is that it brings into focus the crucial role of social influence, particularly the impact of family, friends, and neighbors, on both risk perception and evacuation decisions. Our findings highlight the value of a data-driven approach, which supports and complements theoretical frameworks such as PMT and PADM.

There are several limitations to this work. Firstly, the questionnaire studies were conducted on Amazon Mechanical Turk, which may result in samples that are not representative of the general population in the targeted areas. Additionally, the sample size for the Harvey study is small. Secondly, structure learning algorithms have inherent limitations. Constraint-based algorithms make several assumptions, including faithfulness and the absence of latent confounders, which may not hold in real-world scenarios. Furthermore, these algorithms learn structures that belong to the same I-equivalence class, where some edge directions cannot be distinguished based on conditional independence (CI) tests.

This work is a preliminary step toward building causal inference models for hurricane evacuation decision-making. After consolidating the structure of BNs and estimating model parameters, we can perform causal inference and estimate the causal effects of interventions using do-calculus [16]. For instance, this approach will enable us to answer critical questions, such as determining the causal impact of an official evacuation notice on the decision to evacuate within a specific demographic in a targeted area.

**References**

[1] E. J. Baker, "Hurricane evacuation behavior," International Journal of Mass Emergencies & Disasters, vol. 9, no. 2, pp. 287–310, 1991.
[2] G. Brown, A. Pocock, M.-J. Zhao, and M. Luján, "Conditional likelihood maximisation: A unifying framework for information theoretic feature selection," The journal of machine learning research, vol. 13, no. 1, pp. 27–66, 2012.
[3] D. Colombo, M. H. Maathuis, et al., "Order-independent constraint-based causal structure learning.," J. Mach. Learn. Res., vol. 15, no. 1, pp. 3741–3782, 2014.
[4] A. C. Constantinou, Y. Liu, K. Chobtham, Z. Guo, and N. K. Kitson, "Large-scale empirical validation of Bayesian network structure learning algorithms with noisy data," International Journal of Approximate Reasoning, vol. 131, pp. 151–188, 2021.
[5] N. Dash and H. Gladwin, "Evacuation decision making and behavioral responses: Individual and household," Natural hazards review, vol. 8, no. 3, pp. 69–77, 2007.

[6] N. Fenton and M. Neil, Risk assessment and decision analysis with Bayesian networks. Crc Press, 2018.

[7] S. Hasan, S. Ukkusuri, H. Gladwin, and P. Murray-Tuite, "Behavioral model to understand household-level hurricane evacuation decision making," Journal of Transportation Engineering, vol. 137, no. 5, pp. 341–348, 2011.

[8] S.-K. Huang, M. K. Lindell, and C. S. Prater, "Who leaves and who stays? A review and statistical meta-analysis of hurricane evacuation studies," Environment and Behavior, vol. 48, no. 8, pp. 991–1029, 2016.

[9] S.-K. Huang, M. K. Lindell, and C. S. Prater, "Multistage model of hurricane evacuation decision: Empirical study of hurricanes Katrina and Rita," Natural Hazards Review, vol. 18, no. 3, p. 05 016 008, 2017.

[10] N. K. Kitson, A. C. Constantinou, Z. Guo, Y. Liu, and K. Chobtham, "A survey of Bayesian network structure learning," Artificial Intelligence Review, vol. 56, no. 8, pp. 8721–8814, 2023.

[11] D. Koller and N. Friedman, Probabilistic graphical models: principles and techniques. MIT press, 2009.

[12] M. K. Lindell, J.-C. Lu, and C. S. Prater, "Household decision making and evacuation in response to hurricane Lili," Natural hazards review, vol. 6, no. 4, pp. 171–179, 2005.

[13] M. K. Lindell and R. W. Perry, "The protective action decision model: Theoretical modifications and additional evidence," Risk Analysis: An International Journal, vol. 32, no. 4, pp. 616–632, 2012.

[14] P. Murray-Tuite, W. Yin, S. V. Ukkusuri, and H. Gladwin, "Changes in evacuation decisions between hurricanes Ivan and Katrina," Transportation research record, vol. 2312, no. 1, pp. 98–107, 2012.

[15] NOAA National Centers for Environmental Information (NCEI), U.S. billion-dollar weather and climate disasters, 2024. DOI: 10.25921/stkw-7w73. [Online]. Available: www.ncei.noaa.gov/access/billions/.

[16] J. Pearl, Causality. Cambridge university press, 2009.

[17] R. W. Rogers, "Cognitive and psychological processes in fear appeals and attitude change: A revised theory of protection motivation," Social Psychophysiology: A sourcebook, pp. 153–176, 1983.

[18] K. Sachs, O. Perez, D. Pe'er, D. A. Lauffenburger, and G. P. Nolan, "Causal protein-signaling networks derived from multiparameter single-cell data," Science, vol. 308, no. 5721, pp. 523–529, 2005.

[19] A. R. Sankar, P. Doshi, and A. Goodie, "Evacuate or not? a POMDP model of the decision making of individuals in hurricane evacuation zones," in Uncertainty in Artificial Intelligence, PMLR, 2020, pp. 669–678.

[20] M Scutari, "Learning Bayesian networks with the bnlearn R package," Journal of Statistical Software, vol. 35, no. 3, 2010.

[21] M. Scutari and J.-B. Denis, Bayesian networks: with examples in R. Chapman and Hall/CRC, 2021.

[22] M. Scutari, C. E. Graafland, and J. M. Gutiérrez, "Who learns better Bayesian network structures: Accuracy and speed of structure learning algorithms," International Journal of Approximate Reasoning, vol. 115, pp. 235–253, 2019.

[23] Y. Sun, S.-K. Huang, and X. Zhao, "Predicting hurricane evacuation decisions with interpretable machine learning methods," International Journal of Disaster Risk Science, vol. 15, no. 1, pp. 134–148, 2024.

[24] United States Census Bureau, "United states census data," 2016. [Online]. Available: https://www.census.gov/acs/www/data/data-tables-and-tools/data-profiles/2016/.

[25] S. Yaramakala and D. Margaritis, "Speculative Markov blanket discovery for optimal feature selection," in Fifth IEEE International Conference on Data Mining (ICDM'05), IEEE, 2005, 4–pp.

[26] N. Yongsatianchot and S. Marsella, "A computational model of coping and decision making in high stress, uncertain situations: An application to hurricane evacuation decisions," IEEE Transactions on Affective Computing, 2022.